\documentclass[runningheads]{llncs}
\usepackage{graphicx}
\usepackage{amsmath,amssymb} 
\usepackage{color}
\usepackage[width=122mm,left=12mm,paperwidth=146mm,height=193mm,top=12mm,paperheight=217mm]{geometry}
\begin{document}
\pagestyle{headings}
\mainmatter

\title{Face Detection with End-to-End Integration of a ConvNet and a 3D Model} 

\titlerunning{Face Detection with a ConvNet and a 3D Model}

\authorrunning{Y. Li, B. Sun, T. Wu and Y. Wang}

\author{Yunzhu Li$^{1,*}$, Benyuan Sun$^{1,}$\thanks{Y. Li and B. Sun contributed equally to this work and are joint first authors.}, Tianfu Wu$^{2}$ and  Yizhou Wang$^{1}$ }


\institute{$^1$Nat'l Engineering Laboratory for Video Technology,\\ Key Laboratory of Machine Perception (MoE),\\ 
    Cooperative Medianet Innovation Center, Shanghai\\
	 Sch'l of EECS, Peking University, Beijing, 100871, China\\
	$^2$Department of ECE and the Visual Narrative Cluster,\\ North Carolina State University, Raleigh, USA \\
	\small{\it \{leo.liyunzhu, sunbenyuan, Yizhou.Wang\}@pku.edu.cn, tianfu\_wu@ncsu.edu}}

\maketitle

\begin{abstract}
This paper presents a method for face detection in the wild, which integrates a ConvNet and a 3D mean face model in an end-to-end multi-task discriminative learning framework. The 3D mean face model is predefined and fixed (e.g., we used the one provided in the AFLW dataset~\cite{AFLW}).  The ConvNet consists of two components: (i) The face proposal component computes face bounding box proposals via estimating facial key-points and the 3D transformation (rotation and translation) parameters for each predicted key-point w.r.t. the 3D mean face model.  (ii) The face verification component computes detection results by pruning and refining proposals based on facial key-points based configuration pooling. 
The proposed method addresses two issues in adapting state-of-the-art generic object detection ConvNets (e.g., faster R-CNN~\cite{FasterRCNN}) for face detection: (i) One is to eliminate the heuristic design of predefined anchor boxes in the region proposals network (RPN) by exploiting a 3D mean face model. (ii) The other is to replace the generic RoI (Region-of-Interest) pooling layer with a configuration pooling layer to respect underlying object structures.
The multi-task loss consists of three terms: the classification Softmax loss and the location smooth $l_1$-losses \cite{FastRCNN} of both the facial key-points and the face bounding boxes.
In experiments,
our ConvNet is trained on the AFLW dataset~\cite{AFLW} only and tested on the  FDDB benchmark \cite{FDDB} with fine-tuning and on the AFW benchmark~\cite{AFW} without fine-tuning. The proposed method obtains very competitive state-of-the-art performance in the two benchmarks. 
\end{abstract}

\section{Introduction}
\subsection{Motivation and Objective}
Face detection has been used as a core module in a wide spectrum of applications such as surveillance, mobile communication and human-computer interaction. It is arguably one of the most successful applications of computer vision.  Face detection in the wild continues to play an important role in the era of visual big data (e.g.,  images and videos on the web and in social media). However, it remains a challenging problem in computer vision due to the large appearance variations caused by nuisance variabilities including viewpoints, occlusion, facial expression, resolution, illumination and cosmetics, etc.

\begin{figure*} [ht!]
\centering
\includegraphics[width=1.0\textwidth]{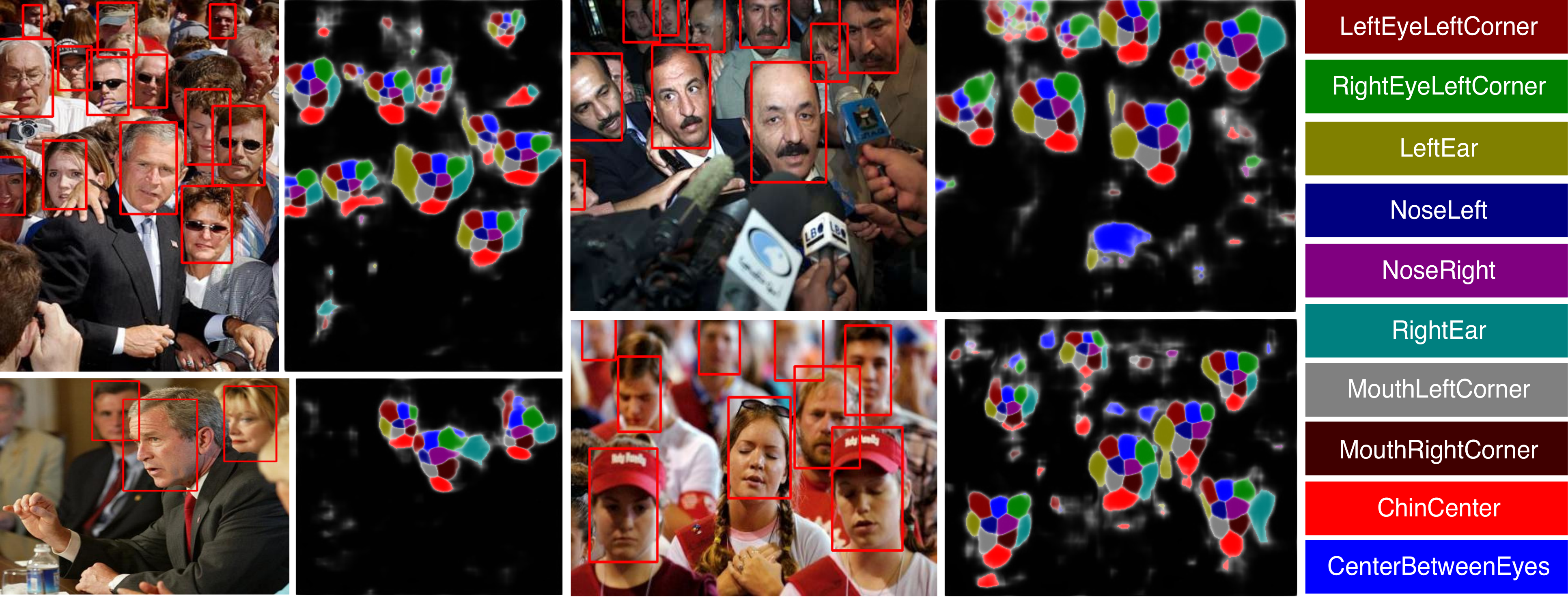}
\caption{Some example results in the FDDB face benchmark~\cite{FDDB} computed by the proposed method. For each testing image, we show the detection results (left) and the corresponding heat map of facial key-points with the legend shown in the right most column. (Best viewed in color) }\label{fig:showcase}
\vspace{-3mm}
\end{figure*}

\begin{figure*}[ht!]
\centering
\includegraphics[width=1.0\textwidth]{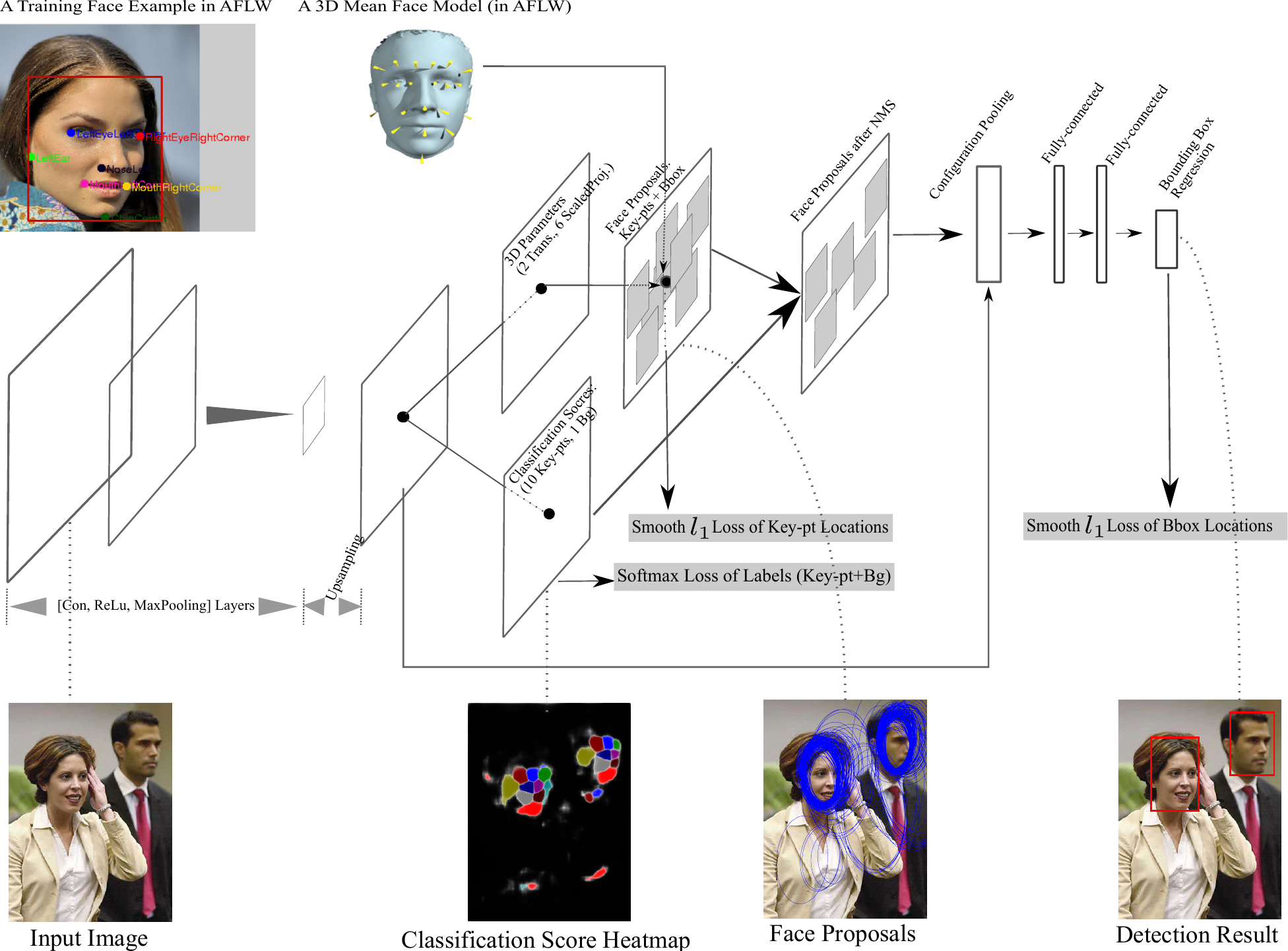}
\caption{Illustration of the proposed method of an end-to-end integration of a ConvNet and a 3D model for face detection (Top), and some intermediate and the final detection results for an input testing image (Bottom). See the legend for the classification score heat map  in Figure~\ref{fig:showcase}. The 3D mean face model is predefined and fixed in both training and testing.  The key idea of the proposed method is to learn a ConvNet to estimate the 3D transformation parameters (rotation and translation) w.r.t. the 3D mean face model to generate accurate face proposals and predict the face key points.  The proposed ConvNet is trained in a multi-task discriminative training framework consisting of the classification Softmax loss and the location smooth $l_1$-losses \cite{FastRCNN} of both the facial key-points and the face bounding boxes. It is surprisingly simple w.r.t. its competitive state-of-the-art performance compared to the other methods in the popular FDDB benchmark~\cite{FDDB} and the AFW benchmark~\cite{AFW}. See text for details.}\label{fig:convnet}
\end{figure*}

It has been a long history that computer vision researchers study how to learn a better representation for unconstrained faces \cite{FaceDetSurvey,Face6Area,Face19Results}. Recently, together with large-scale annotated image datasets such as the ImageNet \cite{ImageNet}, deep ConvNets \cite{LeCunCNN,AlexNet} have made significant progress in  generic object detection \cite{FastRCNN,FasterRCNN,ResidualNet}, as well as in face detection~\cite{FaceCascadeCNN,DP2MFD}. The success is generally considered to be due to the region proposal methods and region-based ConvNets (R-CNN) \cite{RCNN}. The two factors used to be addressed separately (e.g., the popular combination of the Selective Search \cite{SS} and R-CNNs pretrained on the ImageNet), and now they are integrated through introducing the region proposal networks (RPNs) as done in the faster-RCNN \cite{FasterRCNN} or are merged into a single pipeline for speeding up the detection as done in~\cite{YOLO,liu15ssd}.  In R-CNNs, one key layer is the so-called RoI (Region-of-Interest) pooling layer \cite{FastRCNN}, which divides a valid RoI (e.g., an object bounding box proposal) evenly into a grid with a fixed spatial extent (e.g., $7\times 7$) and then uses max-pooling to convert the features inside the RoI into a small feature map. In this paper, we are interested in adapting state-of-the-art ConvNets of generic object detection (e.g., the faster R-CNN~\cite{FasterRCNN}) for face detection by overcoming the following two limitations:
\begin{enumerate}
\item[i)] RPNs need to predefine a number of anchor boxes (with different aspect ratios and sizes), which requires potentially tedious parameter tuning in training and is sensitive to the (unknown) distribution of the aspect ratios and sizes of the object instances in a random testing image.
\item[ii)] The RoI pooling layer in R-CNNs is predefined and generic to all object categories without exploiting the underlying object structural configurations, which either are available from the annotations in the training dataset (e.g., the facial landmark annotations in the AFLW dataset \cite{AFLW}) as done in~\cite{AFW} or can be pursued during learning (such as the deformable part-based models~\cite{DPM,DP2MFD}).
\end{enumerate}

To address the two above issues in learning ConvNets for face detection, we propose to integrate a ConvNet and a 3D mean face model in an end-to-end multi-task discriminative learning framework.
Figure~\ref{fig:showcase} shows some results of the proposed method.

\subsection{Method Overview}
Figure~\ref{fig:convnet} illustrates the proposed method. We use 10 facial key-points in this paper, including {``LeftEyeLeftCorner", ``RightEyeRightCorner",
    ``LeftEar", ``NoseLeft", ``NoseRight", ``RightEar",
    ``MouthLeftCorner",
    ``MouthRightCorner",
    ``ChinCenter",
    ``CenterBetweenEyes"} (see an example image in the left-top of Figure~\ref{fig:convnet}). The 3D mean face model is then represented by the corresponding ten 3D facial key-points. The architecture of our ConvNet is straight-forward when taking into account a 3D model (see Section~\ref{sec:architecture} for details).

 \textbf{The key idea is to learn a ConvNet to (i) estimate the 3D transformation parameters (rotation and translation) w.r.t. the 3D mean face model  for each detected facial key-point so that we can generate face bounding box proposals and (ii) predict facial key-points for each face instance more accurately.}  Leveraging the 3D mean face model is able to ``kill two birds with one stone": Firstly, we can eliminate the manually heuristic design of anchor boxes in RPNs. Secondly, instead of using the generic RoI pooling, we devise a ``configuration pooling" layer so as to respect the object structural configurations in a meaningful and principled way. In other words, we propose to learn to compute the proposals in a straight-forward top-down manner, instead of to design the bottom-up heuristic and then learn related regression parameters. To do so,  we assume a 3D mean face model is available and facial key-points are annotated in the training dataset. Thanks to  many excellent existing work in collecting and annotating face datasets, we can easily obtain both for faces nowadays.  In learning, we have multiple types of losses involved in the objective loss function, including classification Softmax loss and location smooth $l_1$-loss \cite{FastRCNN} of facial key-points, and location smooth $l_1$-loss of face bounding boxes respectively, so we formulate the learning of the proposed ConvNet under the multi-task discriminative deep learning framework (see Section~\ref{sec:training}).

In summary, we provide a clean and straight-forward solution for end-to-end integration of a ConvNet and a 3D model for face detection~\footnote{We use the open source deep learning package, MXNet~\cite{mxnet}, in our implementation. The full source code is released at https://github.com/tfwu/FaceDetection-ConvNet-3D}. In addition to the competitive performance w.r.t the state-of-the-art face detection methods on the FDDB and AFW benchmarks, the proposed method is surprisingly simple and it is able to detect challenging faces (e.g., small, blurry, heavily occluded and extreme poses).

Potentially, the proposed method can be utilized to learn to detect other rigid or semi-rigid object categories (such as cars) if the required information (such as the 3D model and key-point/part annotation) are provided in training.


\section{Related Work}

There are a tremendous amount of existing works on face detection or generic object detection. We refer to \cite{FaceDetSurvey} for a more thorough survey on face detection. We discuss some of the most relevant ones in this section.

In human/animal vision, how  the brain distills a representation of objects from retinal input is one of the central challenges for systems neuroscience, and many works have been focused on  the ecologically important class of objects--faces.   Studies using fMRI experiments in the macaque reveal that faces are represented by a system of six discrete, strongly interconnected regions which illustrates hierarchical information processing in the brain \cite{Face6Area}, as well as some other results~\cite{Face19Results}. These findings provide some biologically-plausible evidences for supporting the usage of deep learning based approaches in face detection and analysis.

The seminal work of Viola and Jones \cite{VJ} made face detection by a computer vision system feasible in real world applications, which trained a cascade of AdaBoost classifiers using Haar wavelet features. Many works followed this direction with different extensions proposed in four aspects: appearance features (beside Haar) including Histogram of Oriented Gradients (HOG)~\cite{HOG},  Aggregate Channel Features (ACF)~\cite{ACF}, Local Binary Pattern (LBP) features~\cite{LBP} and SURF~\cite{SURF}, etc.; detector structures (beside cascade) including the the scalar tree~\cite{ScalarTree} and the width-first-search tree~\cite{MVFD}, etc.; strong classifier learning (beside AdaBoost) including RealBoost~\cite{RealBoost} and GentleBoost~\cite{GentleBoost}, ect; weak classifier learning (beside stump function) including the histogram method~\cite{KLBoost} and the joint binarizations of Haar-like feature~\cite{JointBinHaar}, etc..

Most of the recent face detectors are based on the deformable part-based  model (DPM)~\cite{DPM,AFW,DPMFace} with HOG features used, where a face is represented by a collection of parts defined based on either facial landmarks or heuristic pursuit as done in the original DPM. \cite{DPMFace} showed that a properly trained vanilla DPM can yield significant improvement for face detection.

More recent advances in deep learning~\cite{LeCunCNN,AlexNet} further boosted the face detection performance by learning more discriminative features from large-scale raw data, going beyond those handcrafted ones. In the FDDB benchmark, most of the face detectors with top performance are based on ConvNets \cite{DP2MFD,FaceCascadeCNN}, combining with cascade \cite{FaceCascadeCNN} and more explicit structure \cite{FacePart2Whole}.

3D information has been exploited in learning object models in different ways. Some works~\cite{3DViewBased,3DViewBased2} used a mixture of 3D view based templates by dividing the view sphere into a number of sectors. \cite{3DModel,3DTangram} utilized 3D models in extracting features and inferring the object pose hypothesis based on EM or DP. \cite{DeepFace} used a 3D face model for aligning faces in learning ConvNets for face recognition. Our work resembles \cite{Barbu3DFace} in exploiting 3D model in face detection, which obtained very good performance in the FDDB benchmark. \cite{Barbu3DFace} computes meaningful 3D pose candidates by image-based regression from detected face key-points with traditional handcrafted features, and verifies the 3D pose candidates by a parameter sensitive classifier based on difference features relative to the 3D pose. Our work integrates a ConvNet and a 3D model in an end-to-end multi-task discriminative learning fashion, which is more straightforward and simpler compared to \cite{Barbu3DFace}.

{\bf Our Contributions.}  The proposed method contributes to face detection in three aspects.
\begin{enumerate}
\item[i)] It presents a simple yet effective method to integrate a ConvNet and a 3D model in an end-to-end learning with multi-task loss used for face detection in the wild.
\item[ii)] It addresses two limitations in adapting the state-of-the-art faster RCNN~\cite{FasterRCNN} for face detection: eliminating the heuristic design of anchor boxes by leveraging a 3D model, and replacing the generic and predefined {\em RoI pooling} with a {\em configuration pooling} which exploits the underlying object structural configurations.
\item[iii)] It obtains very competitive state-of-the-art performance in the FDDB~\cite{FDDB} and AFW~\cite{AFW} benchmarks.
\end{enumerate}

\paragraph{Paper Organization.} The remainder of this paper is organized as follows. Section~\ref{sec:method} presents the method of face detection using a 3D model and details of our ConvNet including its architecture and training procedure. Section~\ref{sec:exp} presents details of experimental settings and shows the experimental results in the FDDB and AFW benchmarks. Section~\ref{sec:conclusion} first concludes this paper and then discuss some on-going and future work to extend the proposed work.

\section{The Proposed Method}\label{sec:method}
In this section, we introduce the notations and present details of the proposed method.

\subsection{3D Mean Face Model and Face Representation}\label{sec:3Dparameter}
In this paper, a 3D mean face model is represented by a collection of $n$ 3D key-points in the form of $(x, y, z)$ and then is denoted by a $n\times 3$ matrix, $F^{(3)}$. Usually, each key-point has its own semantic name. We use the 3D mean face model in the AFLW dataset~\cite{AFLW} which consists of 21 key-points. We select 10 key-points as stated above.

A face, denoted by $f$, is presented by its 3D transformation parameters, $\Theta$, for rotation and translation, and a collection of 2D key-points, $F^{(2)}$, in the form of $(x, y)$ (with the number being less than or equal to $n$). Hence, $f=(\Theta, F^{(2)})$. The 3D transformation parameters $\Theta$ are defined by,
\begin{equation}
\Theta = (\mu, s, A^{(3)}),
\end{equation}
where $\mu$ represents a 2D translation $(dx, dy)$, $s$ a scaling factor, and $A^{(3)}$ a $3\times 3$ rotation matrix. We can compute the predicted 2D key-points by,
\begin{equation}
\hat{F}^{(2)} = \mu + s\cdot \pi(A^{(3)} \cdot F^{(3)}),
\end{equation}
where $\pi()$ projects a 3D key-point to a 2D one, that is, $\pi: \mathbb{R}^3 \rightarrow \mathbb{R}^2$ and $\pi(x, y, z) = (x, y)$. Due to the projection $\pi()$, we only need 8 parameters out of the original 12 parameters. Let $A^{(2)}$ denote a $2\times 3$ matrix, which is composed by the top two rows of $A^{(3)}$. We can re-produce the predicted 2D key-points by,
 \begin{equation}
 \hat{F}^{(2)} = \mu + A^{(2)} \cdot F^{(3)}
 \label{eqn:Proj3DPnts}
 \end{equation}
which makes it easy to implement the computation of back-propagation in training our ConvNet.

Note that we use the first sector in a 4-sector $X$-$Y$ coordinate system to define all the positions, that is, the origin point $(0, 0)$ is defined by the left-bottom corner in an image lattice.

In face datasets, faces are usually annotated with bounding boxes. In the FDDB benchmark~\cite{FDDB}, however, faces are annotated with ellipses and detection performance are evaluated based on ellipses.  Given a set of predicted 2D key-points $\hat{F}^{(2)}$, we can compute proposals in both ellipse form and bounding box form. 

\textit{Computing a Face Ellipse and a Face Bounding Box based on a set of Predicted 2D Key-Points.}
For a given $\hat{F}^{(2)}$,  we first predict the position of the top of head by,
\[
\bigl(\begin{smallmatrix}
x \\ y
\end{smallmatrix} \bigr)_\text{TopOfHead} = 2 \times \bigl(\begin{smallmatrix}
x \\ y
\end{smallmatrix} \bigr)_\text{CenterBetweenEyes} - \bigl(\begin{smallmatrix}
x \\ y
\end{smallmatrix} \bigr)_\text{ChinCenter}.
\]
Based on the keypoints of a face proposal, we can compute its ellipse and bounding box. 

\textit{Face Ellipse.} We first compute the outer rectangle. We use as one axis the line segment between the top-of-the-head key-point and the chin key-point, and then compute the minimum rectangle, usually a rotated rectangle, which covers all the key-points. Then, we can compute the ellipse using the two edges of the (rotated) rectangle as the major and minor axes respectively. 

\textit{Face Bounding Box.} We compute a face bounding box by the minimum up-right rectangle which covers all the key-points, which is also adopted in the FDDB benchmark~\cite{FDDB}.

\subsection{The Architecture of Our ConvNet}\label{sec:architecture}
As illustrated in Figure~\ref{fig:convnet}, the architecture of our ConvNet consists of:

 \textit{i) Convolution, ReLu and MaxPooling Layers}. We adopt the VGG~\cite{VGG} design in our experiments which has shown superior performance in a series of tasks. There are 5 groups and each group has 3 convolution and ReLu consecutive layers followed by a MaxPooling layer except for the $5$th group. The spatial extent of the final feature map is of 16 times smaller than that of an input image due to the sub-sampling.

 \textit{ii) An Upsampling Layer.} Since we will measure the location difference between the input facial key-points and the predicted ones, we add an upsampling layer to compensate the sub-sampling effects in previous layers. It is implemented by deconvolution. We upsample the feature maps to 8 times bigger in size (i.e., the upsampled feature maps are still quarter size of an input image) considering the trade-off between key-point location accuracy, memory consumption and computation efficiency.

\textit{iii) A Facial Key-point Label Prediction Layer.} There are 11 labels (10 facial key-points and 1 background class). It is used to compute the classification Softmax loss based on the input in training.

\textit{iv) A 3D Transformation Parameter Estimation Layer.} This is the key observation in this paper. Originally, there are 12 parameters in total consisting of 2D translation, scaling and $3\times 3$ rotation matrix. Since we focus on the 2D projected key-points, we only need to account for 8 parameters (see the derivation above).

 \textit{v) A Face Proposal Layer.} At each position, based on the 3D mean face model and the estimated 3D transformation parameters, we can compute a face proposal consisting of 10 predicted facial key-points and the corresponding face bounding box. \textit{The score of a face proposal} is the sum of $log$ probabilities of the 10 predicted facial key-points. The predicated key-points will be used to compute the smooth $l_1$ loss~\cite{FastRCNN} w.r.t. the ground-truth key-points. We apply the non-maximum suppression (NMS) to the face proposals in which the overlap between two bounding boxes $a$ and $b$ is computed by ${|a|\cap |b|}\over {|b|}$ (where $|\cdot|$ represents the area of a bounding box), instead of the traditional intersection-over-union, accounting for the fact that it is rarely observed that one face is inside another one.

 \textit{vi) A Configuration Pooling Layer.} After NMS, for each face proposal, we pool the features based on the predicted 10 facial key-points. Here, for simplicity, we use all the 10 key-points without considering the invisibilities of certain key-points in different face examples.

\textit{vii) A Face Bounding Box Regression Layer.} It is used to further refine face bounding boxes in the spirit similar to the method~\cite{FastRCNN}. Based on the configuration pooling, we add two fully-connected layers to implement the regression. It is used to compute the smooth $l_1$ loss of face bounding boxes.

Denote by $\omega$ all the parameters in our ConvNet, which will be estimated through multi-task discriminative end-to-end learning.

\subsection{The End-to-End Training}\label{sec:training}
\textbf{Input Data.} Denote by $C=\{0, 1, \cdots, 10\}$ as the key-point labels where $\ell=0$ represents the background class. We use the image-centric sampling trick as done in~\cite{FastRCNN,FasterRCNN}. Without loss of generality, considering a training image with only one face appeared, we have its bounding box, $B=(x, y, w, h)$ and $m$ 2D key-points ($m\leq 10$), $\{(\mathbf{x}_i, \ell_i)_{i=1}^m\}$ where $\mathbf{x}_i = (x_i, y_i)$ is the 2D position of the $i$th key-point and  $\ell_i \geq 1 \in C$. We randomly sample $m$ locations outside the face bounding box $B$ as the background class, $\{(\mathbf{x}_i, \ell_i)_{i=m+1}^{2m}\}$ (where $\ell_i=0,\forall i>m$). Note that in our ConvNet, we use the coordinate of the upsampled feature map which is half size along both axes of the original input. All the key-points and bounding boxes are defined accordingly based on ground-truth annotation.

\textbf{The Classification Softmax Loss of Key-point Labels.} At each position $\mathbf{x}_i$, our ConvNet outputs a discrete probability distribution, $p^{\mathbf{x}_i}=(p^{\mathbf{x}_i}_0, p^{\mathbf{x}_i}_1, \cdots, p^{\mathbf{x}_i}_{10})$, over the 11 classes, which is computed by the Softmax over the 11 scores as usual~\cite{AlexNet}. Then, we have the loss,
\begin{equation}
\mathcal{L}_{cls}(\omega) = -{1\over {2m}} \sum_{i=1}^{2m} \log(p_{\ell_i}^{\mathbf{x}_i})
\end{equation}

\textbf{The Smooth $l_1$ Loss of Key-point Locations.} At each key-point location $\mathbf{x}_i \; (\ell_i \ge 1)$, we compute a face proposal based on the estimated 3D parameters and the 3D mean face, denoted by $\{(\hat{\mathbf{x}}^{(i)}_j, \hat{\ell}_j^{(i)})_{j=1}^{10} \}$ the predicted 10 keypoints. So, for each key-point location $\mathbf{x}_i$, we will have $m$ predicted locations, denoted by $\hat{\mathbf{x}}_{i, j}$ ($j=1, \cdots, m$). 
We follow the definition in~\cite{FastRCNN} to compute the smooth $l_1$ loss for each axis individually.
\begin{align}
\mathcal{L}_{loc}^{pt}(\omega) = {1\over m^2}\sum_{i=1}^m \sum_{j=1}^{m} \sum_{\substack{t\in \{x, y\}}} \text{Smooth}_{l_1}(t_i-\hat{t}_{i,j})
\end{align}
where the smooth term is defined by,
\begin{align}
\text{Smooth}_{l_1}(a) = \begin{cases}
    0.5a^2       & \quad \text{if } |a| < 1\\
    |a| - 0.5  & \quad \text{otherwise}.\\
  \end{cases}
\end{align}

\textbf{Faceness Score.} The faceness score of a face proposal in our ConvNet is computed by the sum of log probabilities of the predicted key-points,
\begin{equation}
    \text{Score}(\hat{\mathbf{x}}_i, \hat{\ell}_i) = \sum_{i=1}^{10} \log (p_{\hat{\ell}_i}^{\hat{\mathbf{x}}_i})
    \label{eqn:Faceness}
\end{equation}
where for simplicity we do not account for the invisibilities of certain key-points. So the current faceness score has the issue of potential double-counting, especially for low-resolution faces. We observed that it hurts the quantitative performance in our experiments. We will address this issue in future work. See some heat maps of key-points in Figure~\ref{fig:showcase}.

\textbf{The Smooth $l_1$ Loss of Bounding Boxes.} For each face bounding box proposal $\hat{B}$ (after NMS), our ConvNet computes its bounding box regression offsets, $t=(t_x, t_y, t_w, t_h)$, where $t$ specifies a scale-invariant translation and log-space height/width shift relative to a proposal, as done in~\cite{FastRCNN,FasterRCNN}. For the ground-truth bounding box $B$, we do the same parameterization and have $v=(v_x, v_y, v_w, v_h)$. Assuming that there are $K$ bounding box proposals, we have,
\begin{align}
\mathcal{L}_{loc}^{box}(\omega) = {1\over K}\sum_{k=1}^K \sum_{\substack{ i\in \{x, y, w, h\}}} \text{Smooth}_{l_1}(t_i-v_i)
\end{align}

So, the overall loss function is defined by,
\begin{equation}
\mathcal{L}(\omega) = \mathcal{L}_{cls}(\omega) + \mathcal{L}_{loc}^{pt}(\omega) + \mathcal{L}_{loc}^{box}(\omega),
\end{equation}
where the third term depends on the output of the first two terms, which makes the loss minimization more challenging.
We adopt a method to implement the differentiable bounding box warping layer, similar to~\cite{ResidualNetSeg}.

\section{Experiments} \label{sec:exp}

In this section, we present the training procedure and implementation details and then show evaluation results on the  FDDB~\cite{FDDB} and AFW~\cite{AFW} benchmarks.

\subsection{Experimental Settings}

\textbf{The Training Dataset.} The only dataset we used for training our model is the AFLW dataset \cite{AFLW}, which contains $25,993$ annotated faces in real-world images. The facial key-points are annotated upon visibility w.r.t. a 3D mean face model with 21 landmarks.  Of the images 70\% are used for training while the remaining is reserved as  a validation set.

\textbf{Training process.} For convenience, the short edge of every image is resized to 600 pixels while preserving the aspect ratio (as done in the faster RCNN~\cite{FasterRCNN}), thus our model learns how to handle faces under various scale. To handle faces of different resolution, we randomly blur images using Gaussian filters in pre-processing. Apart from the rescaling and blurring, no other preprocessing mechanisms (e.g., random crop or left-right flipping) are used. 

We adopt the method of image-centric sampling~\cite{FastRCNN,FasterRCNN} which uses one image at a time in training. Under the consideration that grids around the labeled position share almost the same context information, thus the $3\times3$ grids around every labeled key-point's position are also regarded as the same positive examples, and we randomly choose the same amount of background examples outside the bounding boxes. The convolution filters are initialized by the VGG-16~\cite{VGG} pretrained on the ImageNet~\cite{ImageNet}. We train the network for 13 epoch, and during the process, the learning rate is modified from 0.01 to 0.0001.

\subsection{Evaluation of the intermediate results}
\textbf{Key-points classification in the validation dataset.} As are shown by the heat maps in Figure~\ref{fig:showcase}, our model is capable of detecting facial key-points with rough face configurations preserved, which shows the effectiveness of exploiting the 3D mean face model. Table~\ref{table:ClsResult} shows the key-point classification accuracy  on the validation set in the last epoch in training.

\begin{table} [t!]
\centering
\begin{tabular}{| c | c | c | c |}
 \hline
 Category &  Accuracy & Category &  Accuracy \\ \hline
 Background & 97.94\% & LeftEyeLeftCorner & 99.12\% \\
 RightEyeRightCorner & 94.57\% & LeftEar & 95.50\% \\
 NoseLeft & 98.48\% & NoseRight & 97.78\% \\
 RightEar & 91.44\% & MouthLeftCorner & 97.97\% \\
 MouthRightCorner & 98.64\% & ChinCenter & 98.65\% \\
 CenterBetweenEyes  & 96.04\% & AverageDetectionRate & 97.50\% \\ \hline
\end{tabular}
\vspace{1em}
\caption{Classification accuracy of the key-points in the AFLW validation set at the end training. }
\label{table:ClsResult} \vspace{-3mm}
\end{table}

\begin{figure*} 
\centering
\includegraphics[width=0.22\textwidth]{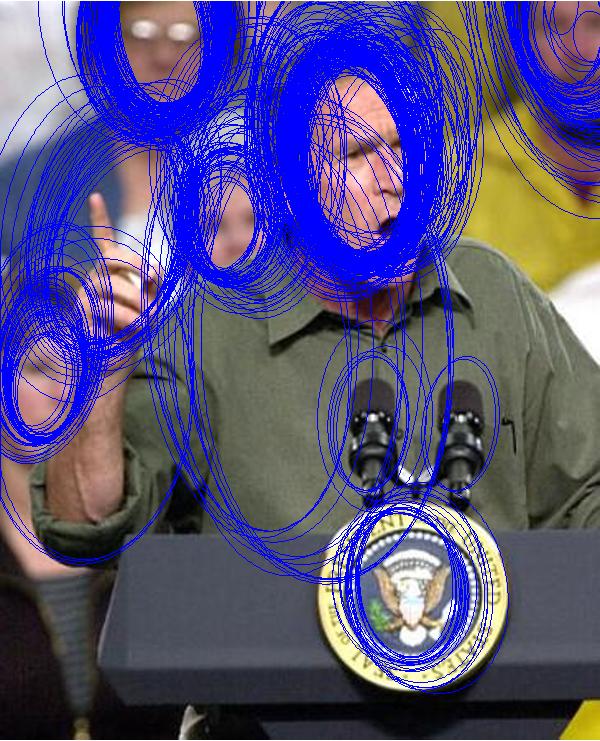}
\includegraphics[width=0.23\textwidth]{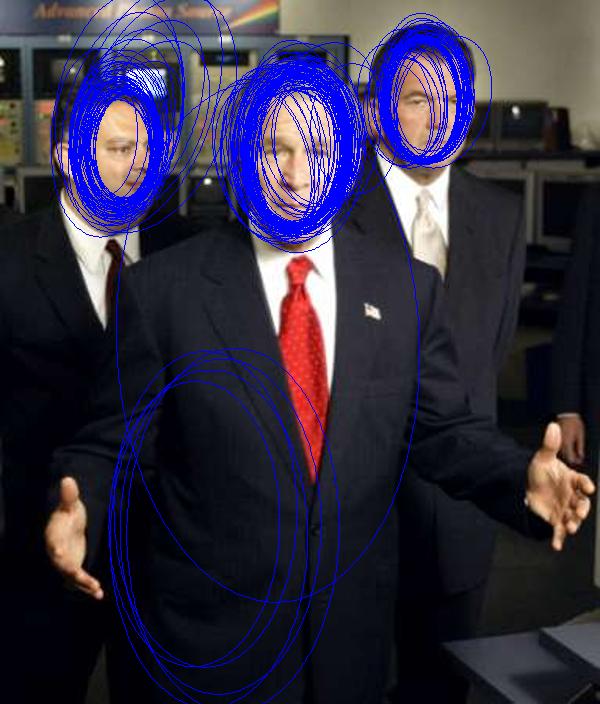}
\includegraphics[width=0.24\textwidth]{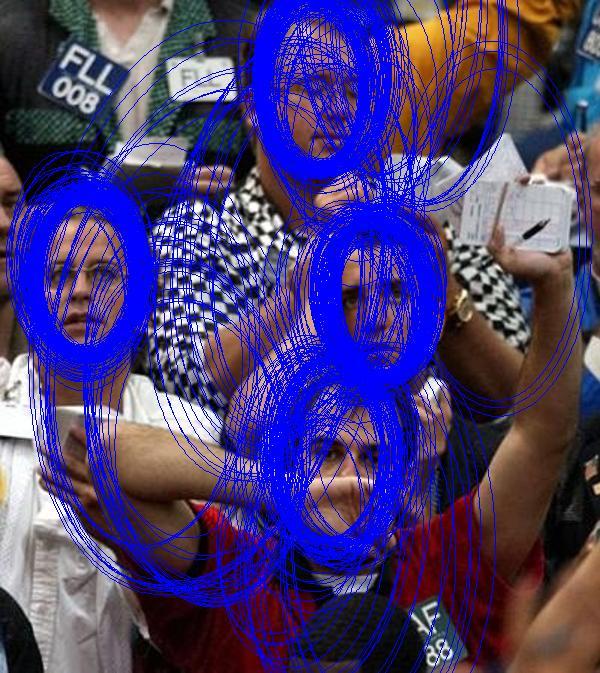}
\includegraphics[width=0.23\textwidth]{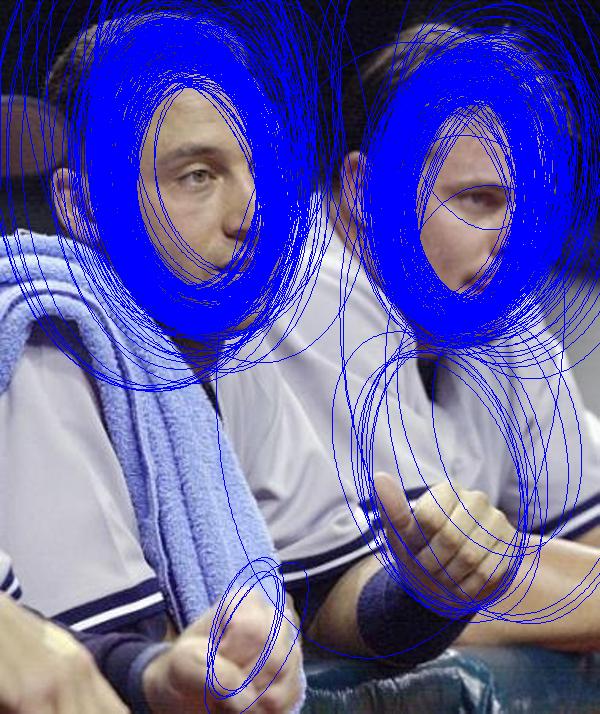}
\caption{Examples of face proposals  computed using predicted 3D transformation parameters without non-maximum suppression. For clarity, we randomly sample 1/30 of the original number of proposals.}\label{fig:FaceProposal}
\end{figure*}

\textbf{Face proposals.} To evaluate the quality of our face proposals, we first show some qualitative results on the FDDB dataset in Fig.~\ref{fig:FaceProposal}. These ellipses are directly calculated from the predicted 3D transformation parameters,  forming several clusters around  face instances. We also evaluate the quantitative results of face proposals.  After a non-maximum suppression of IoU 0.7,  the recall rate of 93.67\% is obtained with  average 34.4 proposals per image.

\begin{figure*}[h!]
\centering
\includegraphics[width=0.95\textwidth]{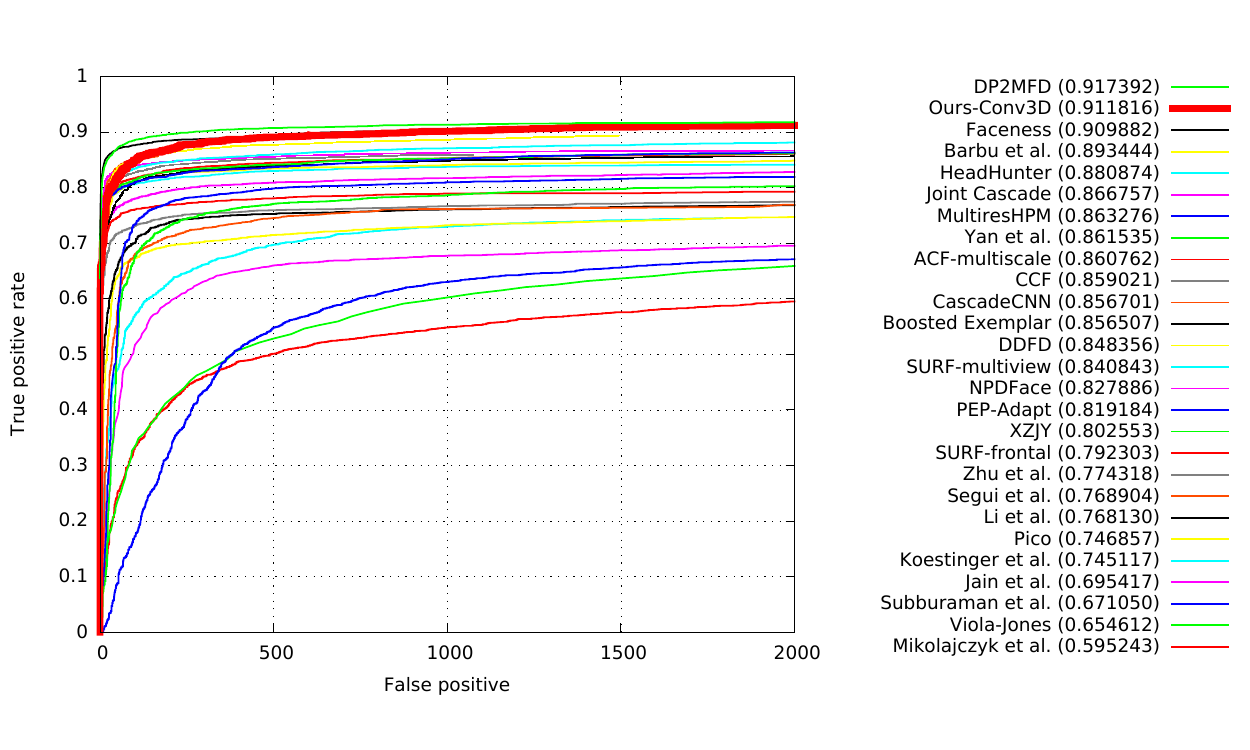}
\caption{FDDB results based on discrete scores using face bounding boxes in evaluation. The recall rates are computed against 2000 false positives.}\label{fig:fddb-discROC}\vspace{-3mm}
\end{figure*}

\begin{figure*}[h!]
	\centering
	\includegraphics[width=0.95\textwidth]{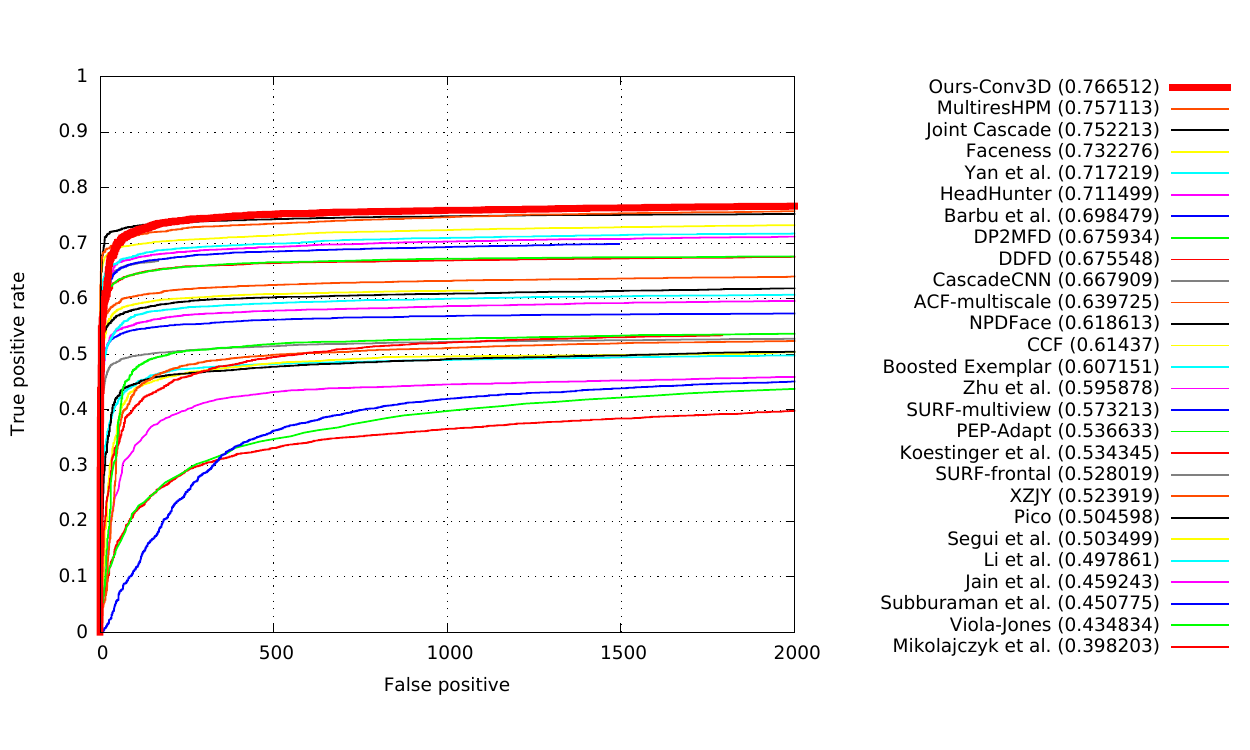}
	\caption{FDDB results based on continuous scores using face ellipses in evaluation. The recall rates are computed against 2000 false positives.}\label{fig:fddb-contROC}
\end{figure*}

\subsection{Face Detection Results}

To show the effectiveness of our method, we test our model on two popular face detection benchmarks: FDDB~\cite{FDDB} and AFW~\cite{AFW}.

\textbf{Results on FDDB.} FDDB is a challenge benchmark for face detection in unconstrained environment, which contains the annotations for 5171 faces in a set of 2845 images. We evaluate our results by using the evaluation code provided by the FDDB authors. The results on the FDDB dataset are shown in Figure~\ref{fig:fddb-discROC}. Our result is represented by "Ours-Conv3D", which surpasses the recall rate of 90\% when encountering 2000 false positives and is competitive to the state-of-the-art methods. We compare with published methods only. Only DP2MFD~\cite{DP2MFD} is slightly better than our model on discrete scores. It's worth noting that we beat all other methods on continuous scores. This is partly caused by the predefined 3D face model helps us better describe the pose and part locations of faces.  We refer to the FDDB result webpage\footnote{http://vis-www.cs.umass.edu/fddb/results.html} for details of the published methods evaluated on it (Fig.~\ref{fig:fddb-discROC} and Fig.~\ref{fig:fddb-contROC}).

When comparing with recent work Faceness~\cite{FacePart2Whole}, we both recognize that one of the central issues to alleviate the problems of the occlusion and pose variation is to introduce facial part detector. However, our mechanism of computing face bounding box candidates is more straight forward since we explicitly integrate the structural information of a 3D mean face model instead of using a heuristic way of assuming the facial part distribution over a bounding box. 

\begin{figure} [t]
\centering
\includegraphics[width=0.9\textwidth]{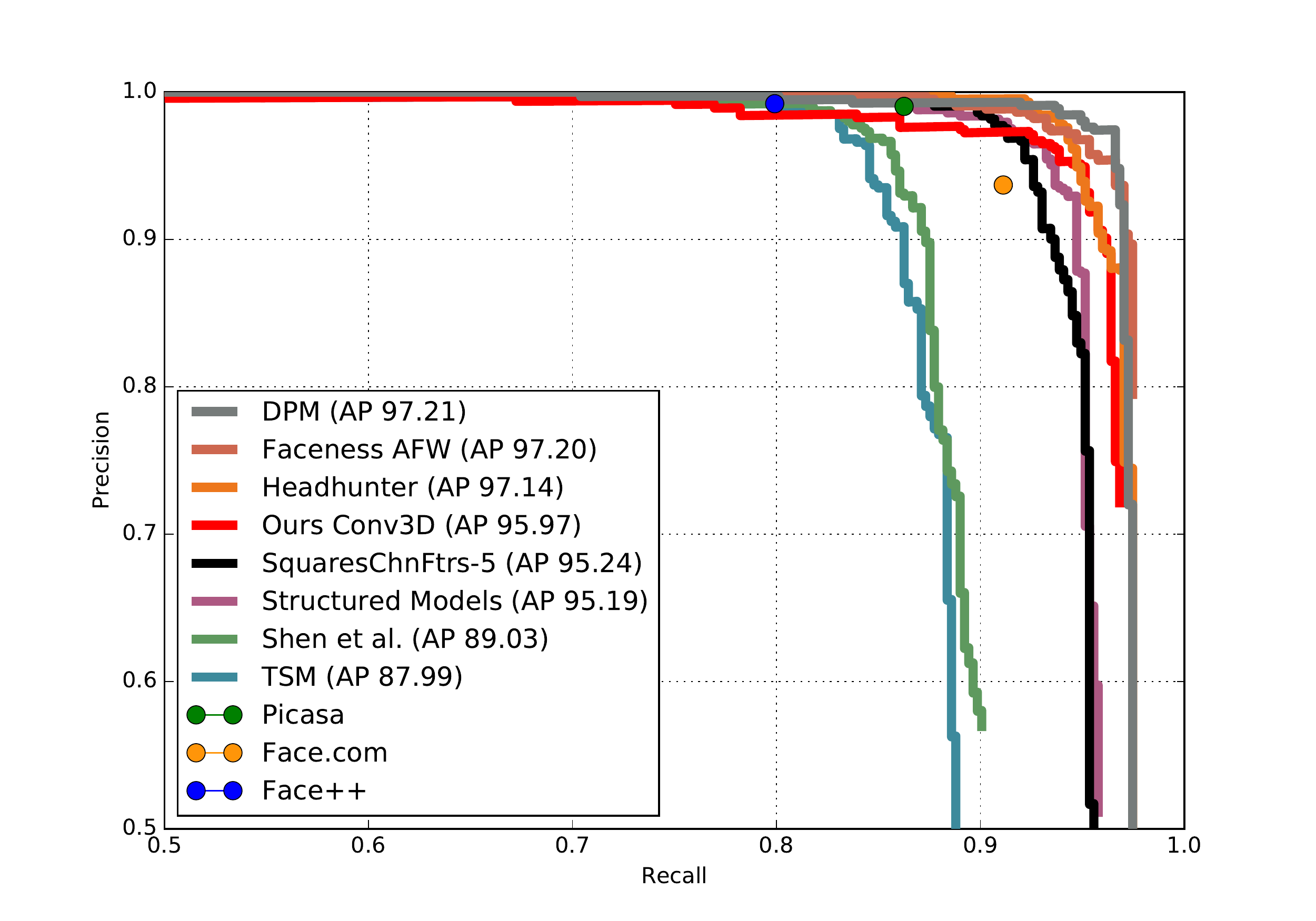}
\caption{Precision-recall curves on the AFW dataset (AP = average precision) without configuration pool and face bounding box regression used.}\label{fig:afw-ROC}
\end{figure}

\textbf{Results on AFW.} AFW dataset contains 205 images with faces in various poses and view points. We use the evaluation toolbox provided by \cite{DPMFace}, which contains updated annotations for the AFW dataset where the original annotations are not comprehensive enough. Since the method of labeling face bounding boxes in AFW is different from that of in FDDB, we only use face proposals without configuration pooling and bounding box regression. The results on AFW  are shown in Figure~\ref{fig:afw-ROC}.

\begin{figure*} 
    \centering
    \includegraphics[width=0.3\textwidth]{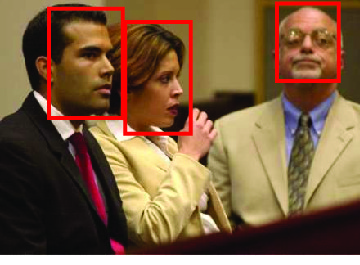}
	\includegraphics[width=0.3\textwidth] {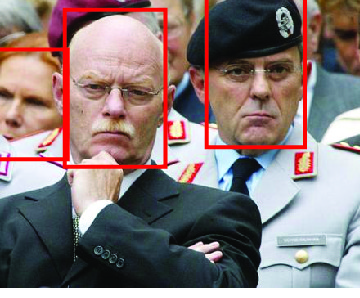}
	\includegraphics[width=0.3\textwidth] {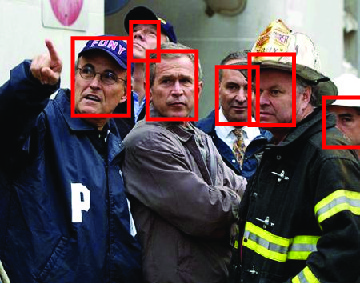} \\
	\includegraphics[width=0.3\textwidth] {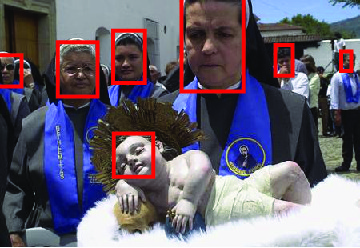}
	\includegraphics[width=0.3\textwidth] {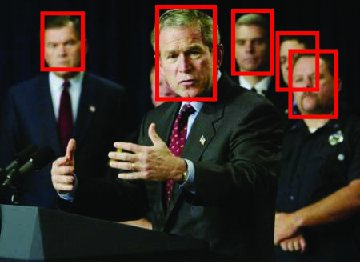}
	\includegraphics[width=0.3\textwidth] {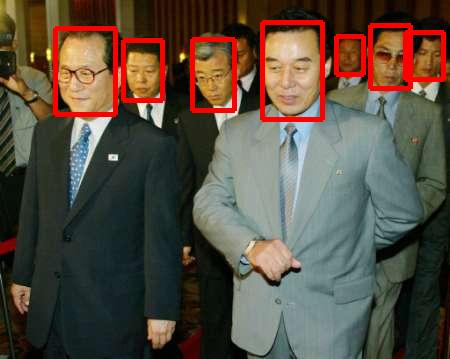}
    \caption{Some qualitative results on the FDDB dataset}
    \label{fig:fddb-showcase} \vspace{-4mm}
\end{figure*}

\begin{figure*} 
    \centering
    \includegraphics[width=0.3\textwidth]{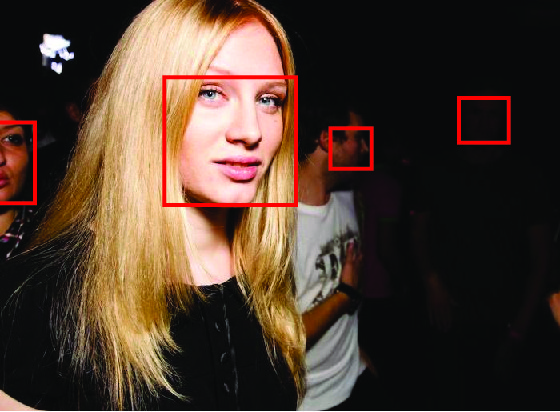}
	\includegraphics[width=0.3\textwidth] {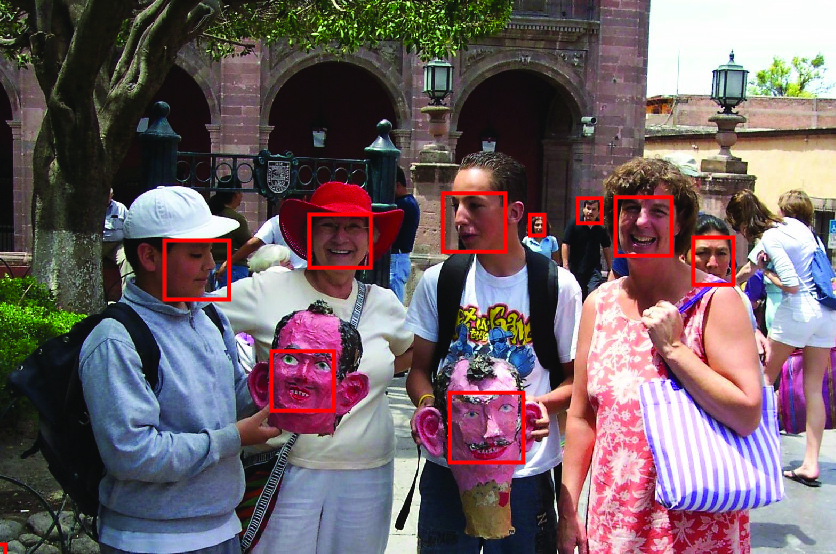}
	\includegraphics[width=0.3\textwidth] {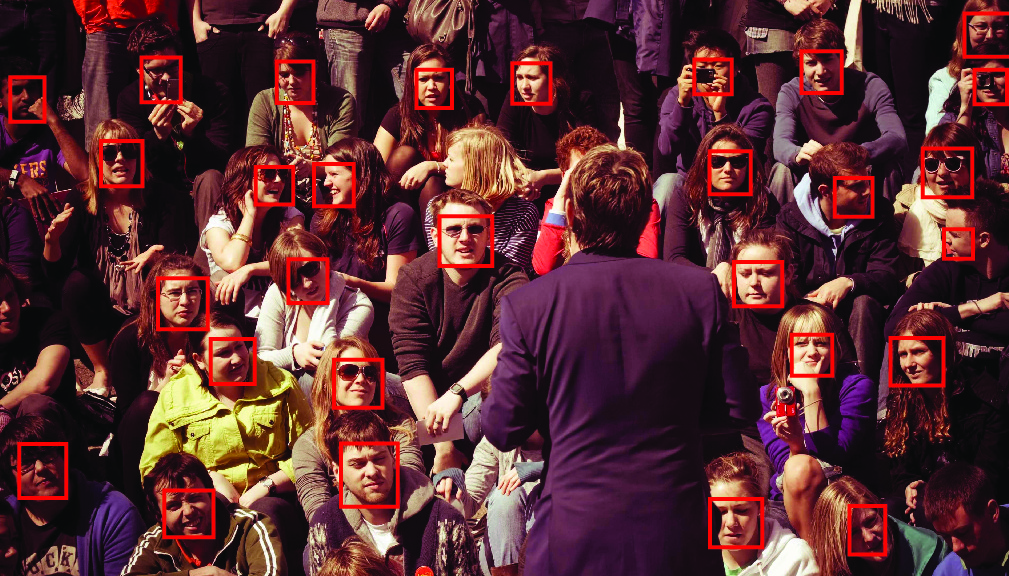}\\
	\includegraphics[width=0.3\textwidth] {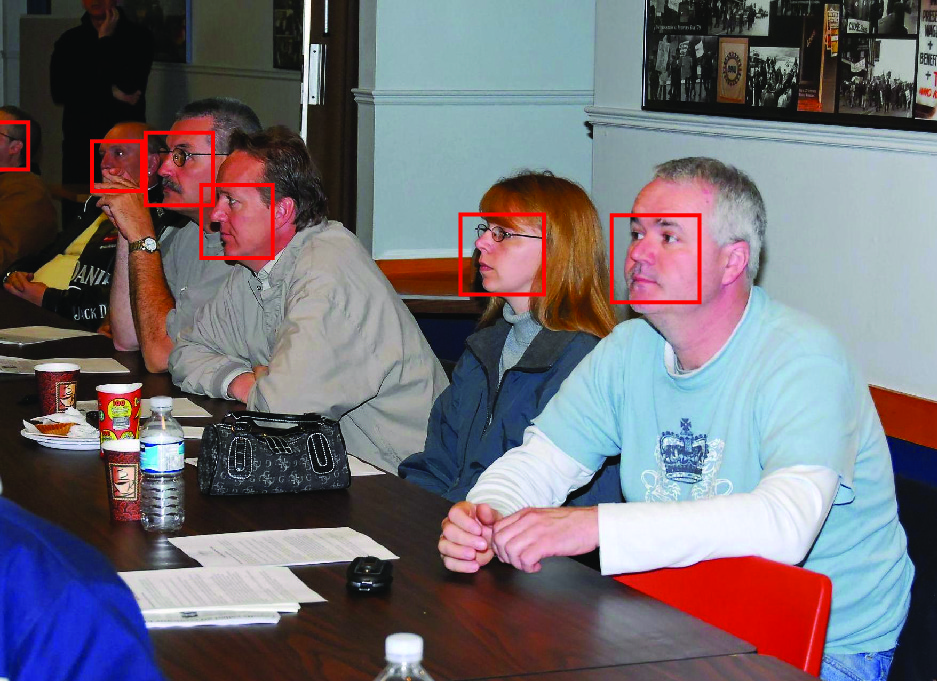}
	\includegraphics[width=0.3\textwidth] {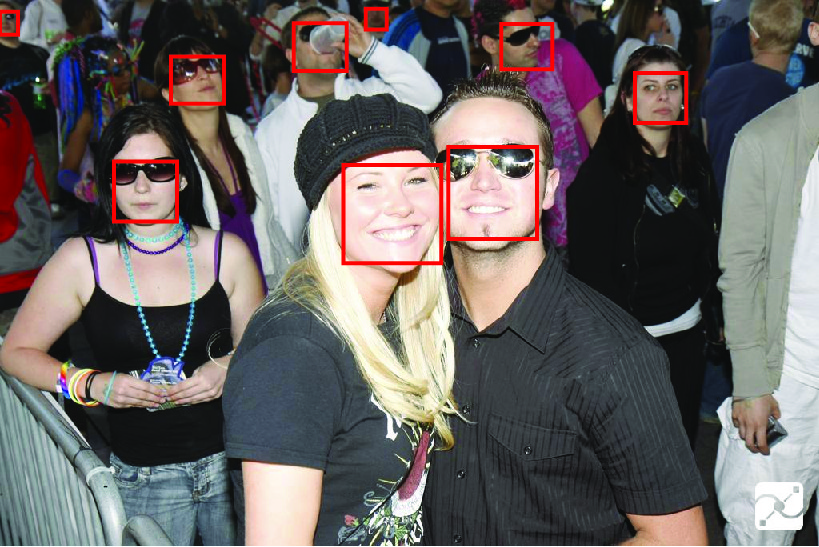}
	\includegraphics[width=0.3\textwidth] {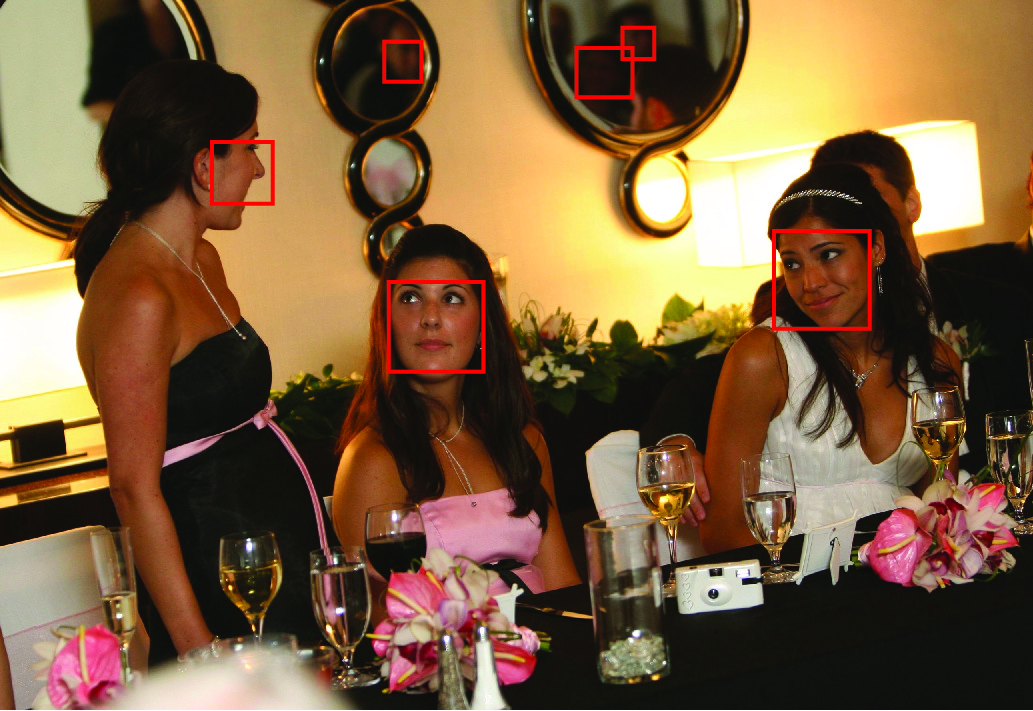}\\
	\includegraphics[width=0.3\textwidth] {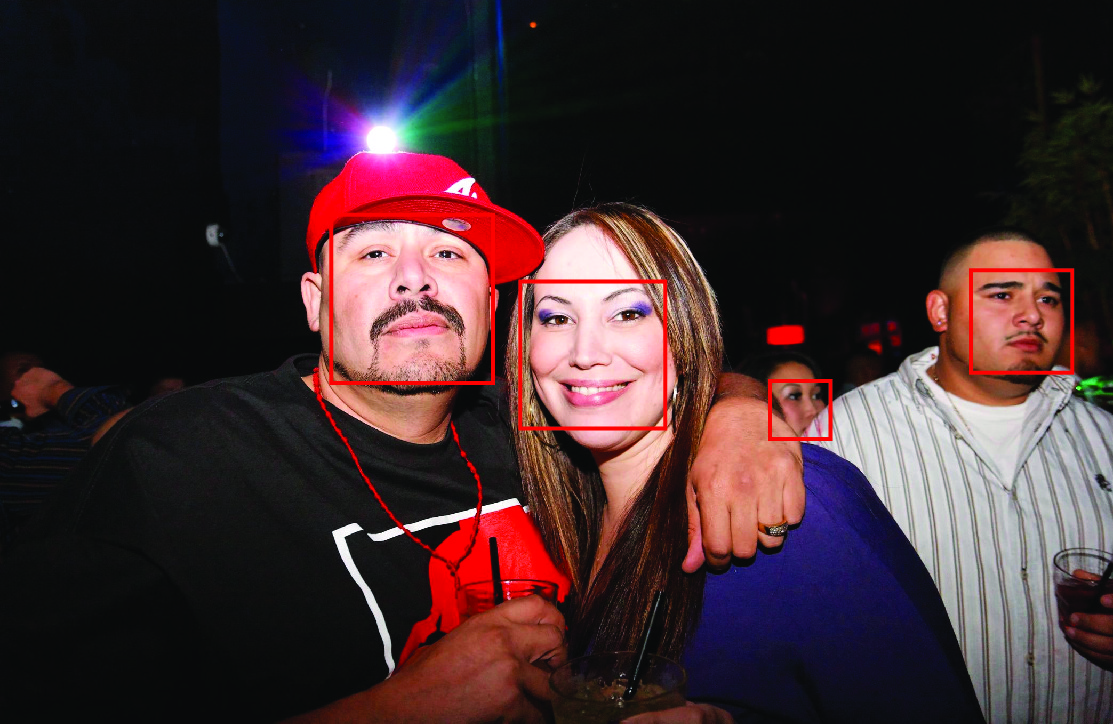}
	\includegraphics[width=0.3\textwidth] {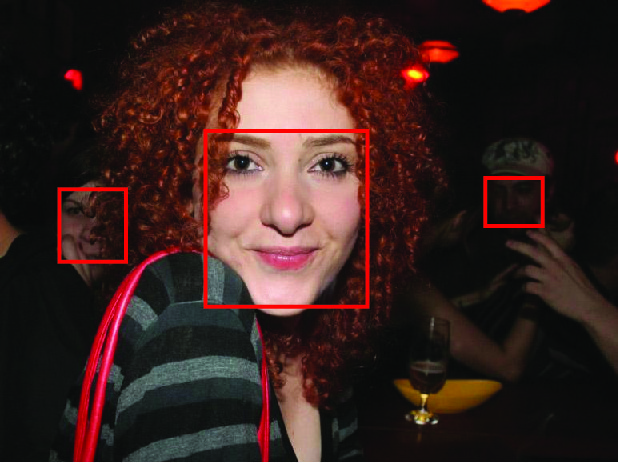}
	\includegraphics[width=0.3\textwidth] {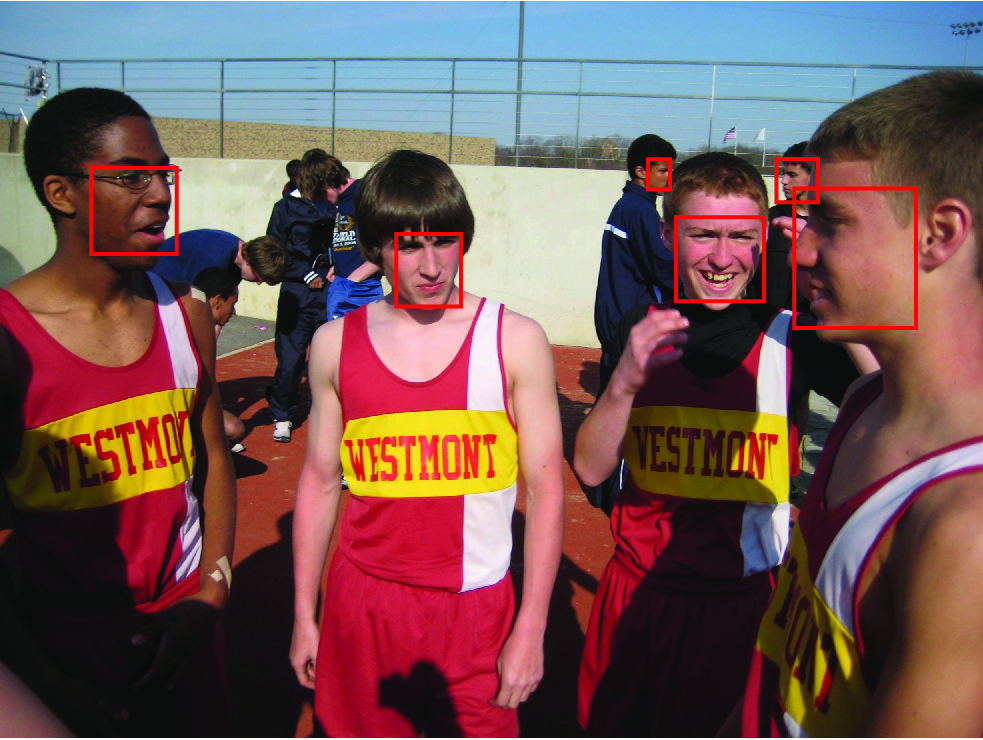}
    \caption{Some qualitative results on the AFW dataset}
    \label{fig:afw-showcase}
\end{figure*}

In our current implementation, there is one major limitation that prevents us from achieving better results.  We do not explicitly handle invisible facial parts, which would be harmful when calculating the faceness score according to Eqn.~\ref{eqn:Faceness}, we will refine the method and introduce mechanisms of handling the invisible problem in future work. More detection results on both datasets are shown in Figure~\ref{fig:fddb-showcase} and Figure~\ref{fig:afw-showcase}.

\section{Conclusion and Discussion} \label{sec:conclusion}
We have presented a method of end-to-end integration of a ConvNet and a 3D model for face detection in the wild. Our method is a clean and straightforward solution when taking into account a 3D model in face detection. It also addresses two issues in state-of-the-art generic object detection ConvNets: eliminating heuristic design of anchor boxes by leveraging a 3D model, and overcoming generic and predefined RoI pooling by configuration pooling which exploits underlying object configurations. In experiments, we tested our method on two benchmarks, the FDDB dataset and the AFW dataset, with very compatible state-of-the-art performance obtained. We analyzed the experimental results and pointed out some current limitations.

In our on-going work, we are working on addressing the doubling-counting issue of the faceness score in the current implementation. We are also working on extending the proposed method for other types of rigid/semi-rigid object classes (e.g., cars). We expect that we will have a unified model for cars and faces which can achieve state-of-the-art performance, which will be very useful in a lot of practical applications such as surveillance and driveless cars.

\textbf{Acknowledgement.}  Y. Li, B. Sun and Y. Wang were supported in part by China 973 Program under Grant no. 2015CB351800, and NSFC-61231010, 61527804, 61421062, 61210005. T. Wu was supported by the ECE startup fund  201473-02119 at NCSU. T. Wu also gratefully acknowledge the support of NVIDIA Corporation with the donation of one GPU.

\bibliographystyle{splncs03}
\bibliography{egbib}

\begin{thebibliography}{10}
\providecommand{\url}[1]{\texttt{#1}}
\providecommand{\urlprefix}{URL }

\bibitem{LBP}
Ahonen, T., Hadid, A., Pietik{\"{a}}inen, M.: Face description with local
  binary patterns: Application to face recognition. {IEEE} Trans. Pattern Anal.
  Mach. Intell.  28(12),  2037--2041 (2006)

\bibitem{Barbu3DFace}
Barbu, A., Gramajo, G.: Face detection using a 3d model on face keypoints. CoRR
   abs/1404.3596 (2014)

\bibitem{SURF}
Bay, H., Ess, A., Tuytelaars, T., Gool, L.J.V.: Speeded-up robust features
  {(SURF)}. Computer Vision and Image Understanding  110(3),  346--359 (2008)

\bibitem{VGG}
Chatfield, K., Simonyan, K., Vedaldi, A., Zisserman, A.: Return of the devil in
  the details: Delving deep into convolutional nets. In: British Machine Vision
  Conference (2014)

\bibitem{mxnet}
Chen, T., Li, M., Li, Y., Lin, M., Wang, N., Wang, M., Xiao, T., Xu, B., Zhang,
  C., Zhang, Z.: Mxnet: {A} flexible and efficient machine learning library for
  heterogeneous distributed systems. CoRR  abs/1512.01274 (2015)

\bibitem{ResidualNetSeg}
Dai, J., He, K., Sun, J.: Instance-aware semantic segmentation via multi-task
  network cascades. In: CVPR (2016)

\bibitem{HOG}
Dalal, N., Triggs, B.: Histograms of oriented gradients for human detection.
  In: CVPR (2005)

\bibitem{ImageNet}
Deng, J., Dong, W., Socher, R., Li, L., Li, K., Li, F.: Imagenet: {A}
  large-scale hierarchical image database. In: CVPR. pp. 248--255 (2009)

\bibitem{ACF}
Doll{\'{a}}r, P., Appel, R., Belongie, S.J., Perona, P.: Fast feature pyramids
  for object detection. {IEEE} Trans. Pattern Anal. Mach. Intell.  36(8),
  1532--1545 (2014)

\bibitem{DPM}
Felzenszwalb, P.F., Girshick, R.B., McAllester, D.A., Ramanan, D.: Object
  detection with discriminatively trained part-based models. {IEEE} Trans.
  Pattern Anal. Mach. Intell.  32(9),  1627--1645 (2010)

\bibitem{ScalarTree}
Fleuret, F., Geman, D.: Coarse-to-fine face detection. International Journal of
  Computer Vision  41(1/2),  85--107 (2001)

\bibitem{Face6Area}
Freiwald, W.A., Tsao, D.Y.: Functional compartmentalization and viewpoint
  generalization within the macaque face-processing system. Science  330(6005),
   845--851 (2010)

\bibitem{GentleBoost}
Friedman, J., Hastie, T., Tibshirani, R.: Additive logistic regression: a
  statistical view of boosting. Annals of Statistics  28,  2000 (1998)

\bibitem{FastRCNN}
Girshick, R.: Fast {R-CNN}. In: ICCV (2015)

\bibitem{RCNN}
Girshick, R.B., Donahue, J., Darrell, T., Malik, J.: Region-based convolutional
  networks for accurate object detection and segmentation. {IEEE} Trans.
  Pattern Anal. Mach. Intell.  38(1),  142--158 (2016)

\bibitem{ResidualNet}
He, K., Zhang, X., Ren, S., Sun, J.: Deep residual learning for image
  recognition. In: CVPR (2016)

\bibitem{3DTangram}
Hu, W., Zhu, S.: Learning 3d object templates by quantizing geometry and
  appearance spaces. {IEEE} Trans. Pattern Anal. Mach. Intell.  37(6),
  1190--1205 (2015)

\bibitem{MVFD}
Huang, C., Ai, H., Li, Y., Lao, S.: High-performance rotation invariant
  multiview face detection. {IEEE} Trans. Pattern Anal. Mach. Intell.  29(4),
  671--686 (2007)

\bibitem{FDDB}
Jain, V., Learned-Miller, E.: Fddb: A benchmark for face detection in
  unconstrained settings. Tech. Rep. UM-CS-2010-009, University of
  Massachusetts, Amherst (2010)

\bibitem{AFLW}
Koestinger, M., Wohlhart, P., Roth, P.M., Bischof, H.: Annotated facial
  landmarks in the wild: A large-scale, real-world database for facial landmark
  localization. In: First IEEE International Workshop on Benchmarking Facial
  Image Analysis Technologies (2011)

\bibitem{AlexNet}
Krizhevsky, A., Sutskever, I., Hinton, G.E.: Imagenet classification with deep
  convolutional neural networks. In: NIPS (2012)

\bibitem{LeCunCNN}
LeCun, Y., Bottou, L., Bengio, Y., Haffner, P.: Gradient-based learning applied
  to document recognition. Proceedings of the IEEE  86(11),  2278--2324 (1998)

\bibitem{FaceCascadeCNN}
Li, H., Lin, Z., Shen, X., Brandt, J., Hua, G.: A convolutional neural network
  cascade for face detection. In: CVPR (2015)

\bibitem{3DModel}
Liebelt, J., Schmid, C.: Multi-view object class detection with a 3d geometric
  model. In: CVPR (2010)

\bibitem{KLBoost}
Liu, C., Shum, H.: Kullback-leibler boosting. In: CVPR (2003)

\bibitem{liu15ssd}
Liu, W., Anguelov, D., Erhan, D., Szegedy, C., Reed, S., Fu, C.Y., Berg, A.C.:
  {SSD}: Single shot multibox detector. arXiv preprint arXiv:1512.02325  (2015)

\bibitem{DPMFace}
Mathias, M., Benenson, R., Pedersoli, M., Gool, L.V.: Face detection without
  bells and whistles. In: ECCV (2014)

\bibitem{JointBinHaar}
Mita, T., Kaneko, T., Hori, O.: Joint haar-like features for face detection.
  In: ICCV (2005)

\bibitem{3DViewBased}
Payet, N., Todorovic, S.: From contours to 3d object detection and pose
  estimation. In: ICCV (2011)

\bibitem{DP2MFD}
Ranjan, R., Patel, V.M., Chellappa, R.: A deep pyramid deformable part model
  for face detection. In: {IEEE} 7th International Conference on Biometrics
  Theory, Applications and Systems (2015)

\bibitem{YOLO}
Redmon, J., Divvala, S., Girshick, R., Farhadi, A.: You only look once:
  Unified, real-time object detection. In: CVPR (2016)

\bibitem{FasterRCNN}
Ren, S., He, K., Girshick, R., Sun, J.: Faster {R-CNN}: Towards real-time
  object detection with region proposal networks. In: NIPS (2015)

\bibitem{RealBoost}
Schapire, R.E., Singer, Y.: Improved boosting algorithms using confidence-rated
  predictions. Machine Learning  37(3),  297--336 (1999)

\bibitem{Face19Results}
Sinha, P., Balas, B., Ostrovsky, Y., Russell, R.: Face recognition by humans:
  19 results all computer vision researchers should know about. Proceedings of
  the IEEE  94(11),  1948--1962 (2006)

\bibitem{3DViewBased2}
Su, H., Sun, M., Li, F., Savarese, S.: Learning a dense multi-view
  representation for detection, viewpoint classification and synthesis of
  object categories. In: ICCV (2009)

\bibitem{DeepFace}
Taigman, Y., Yang, M., Ranzato, M., Wolf, L.: Deepface: Closing the gap to
  human-level performance in face verification. In: CVPR (2014)

\bibitem{SS}
Uijlings, J.R.R., van~de Sande, K.E.A., Gevers, T., Smeulders, A.W.M.:
  Selective search for object recognition. International Journal of Computer
  Vision  104(2),  154--171 (2013)

\bibitem{VJ}
Viola, P.A., Jones, M.J.: Robust real-time face detection. International
  Journal of Computer Vision  57(2),  137--154 (2004)

\bibitem{FacePart2Whole}
Yang, S., Luo, P., Loy, C.C., Tang, X.: From facial parts responses to face
  detection: {A} deep learning approach. In: ICCV (2015)

\bibitem{FaceDetSurvey}
Zafeiriou, S., Zhang, C., Zhang, Z.: A survey on face detection in the wild.
  Comput. Vis. Image Underst.  138(C),  1--24 (Sep 2015)

\bibitem{AFW}
Zhu, X., Ramanan, D.: Face detection, pose estimation, and landmark
  localization in the wild. In: CVPR (2012)

\end{thebibliography}
\end{document}